\documentclass[11pt,a4paper]{article}
\usepackage[hyperref]{naaclhlt2019}
\usepackage{times}
\usepackage{latexsym}
\usepackage{multirow, makecell}
\usepackage{rotating}
\usepackage{amsmath}
\usepackage{colortbl}
\usepackage{diagbox}
\usepackage{subcaption}
\usepackage{url}
\usepackage{textcomp}
\usepackage[dvipsnames]{xcolor}
\usepackage{soul}
\usepackage{booktabs}
\usepackage{balance}

\aclfinalcopy % Uncomment this line for the final submission
\def\aclpaperid{1541} %  Enter the acl Paper ID here

\title{Outlier Detection for Improved Data Quality and Diversity in Dialog Systems}

\author{Stefan Larson\hspace{20pt} Anish Mahendran\hspace{20pt} Andrew Lee\hspace{20pt} Jonathan K. Kummerfeld\\
\textbf{Parker Hill\hspace{8pt} Michael A. Laurenzano\hspace{8pt} Johann Hauswald\hspace{8pt} Lingjia Tang\hspace{8pt} Jason Mars}\\
  Clinc, Inc. \\
  Ann Arbor, MI, USA \\
  \texttt{\{stefan,anish,andrew,jkk\}@clinc.com}\\
  \texttt{\{parkerhh,mike,johann,lingjia,jason\}@clinc.com}
  }

\date{}

\begin{document}
\maketitle
\begin{abstract}
In a corpus of data, outliers are either \emph{errors}: mistakes in the data that are counterproductive, or are \emph{unique}: informative samples that improve model robustness.
Identifying outliers can lead to better datasets by (1) removing noise in datasets and (2) guiding collection of additional data to fill gaps.
However, the problem of detecting both outlier types has received relatively little attention in NLP, particularly for dialog systems.
We introduce a simple and effective technique for detecting both erroneous and unique samples in a corpus of short texts using neural sentence embeddings combined with distance-based outlier detection.
We also present a novel data collection pipeline built atop our detection technique to automatically and iteratively mine unique data samples while discarding erroneous samples.
Experiments show that our outlier detection technique is effective at finding errors while our data collection pipeline yields highly diverse corpora that in turn produce more robust intent classification and slot-filling models.
\end{abstract}

\section{Introduction}

High-quality annotated data is one of the fundamental drivers of progress in Natural Language Processing \citep[e.g.][]{ptb,europarl}. %ud
In order to be effective at producing an accurate and robust model, a dataset needs to be correct while also diverse enough to cover the full range of ways in which the phenomena it targets occur.
Substantial research effort has considered dataset correctness \citep{Eskin:2000,Markus:2003,Rehbein:2017}, particularly for crowdsourcing \citep{Snow:2008,Jiang:2017},
but addressing diversity in data has received less attention, with the exception of using data from diverse domains \citep{ontonotes}.
Outlier detection, the task of finding examples in a dataset that are atypical, provides a means of approaching the questions of correctness and diversity, but has mainly been studied at the document level \citep{Guthrie:2008,Zhuang:2017},
whereas texts in dialog systems are often no more than a few sentences in length.

We propose a novel approach that uses sentence embeddings to detect outliers in a corpus of short texts.
We rank samples based on their distance from the mean embedding of the corpus and consider samples farthest from the mean outliers.
Outliers come in two varieties:
(1) \emph{errors}, sentences that have been mislabeled whose inclusion in the dataset would be detrimental to model performance,
and (2) \emph{unique} samples, sentences that differ in structure or content from most in the data and whose inclusion would be helpful for model robustness.
Building upon this approach, we propose a novel crowdsourcing pipeline that distinguishes errors from unique samples and uses the unique samples to guide workers to give more diverse samples.

Experimentally, we find that our outlier detection technique leads to efficient detection of both artificial and real errors in our datasets.
We also use the proposed crowdsourcing pipeline to collect new datasets and build models for the dialog system tasks of intent classification and slot-filling.
We find that the proposed pipeline produces more diverse data, which in turn results in models that are more robust.

\section{Background and Related Work}

\subsection{Outlier Detection}
Outlier detection \citep{Rousseeuw:1987}, also called outlier analysis \citep{Aggarwal:2015} or anomaly detection \cite{Chandola:2009}, is the task of identifying examples in a dataset that differ substantially from the rest of the data.

For almost two decades, a body of work in NLP has investigated applying these ideas to data in order to identify annotation errors \citep{Abney:1999}.
Approaches have included the use of scores from trained models for POS tagging \citep{Abney:1999,Eskin:2000,W00-1907,W11-0408,Fukumoto:2004}, count-based methods that compare examples from across the corpus \citep{C02-1101,HOLLENSTEIN16.584}, characterizing data based on feature vectors projected down into a low-dimensional space \citep{Guthrie:2008}, and tracking the difficulty of learning each example during training \citep{Amiri:2018}.
One particularly effective approach has been to find \textit{n}-grams that match but have different labels, as shown for annotations including POS tags \citep{Markus:2003}, syntactic parses \citep{P05-1040,P10-1075,W11-2929}, and predicate-argument relations \citep{L08-1162}.
Our work instead uses continuous representations of text derived from neural networks.

While finding errors is an extremely useful application of outlier detection, also of interest are examples that are correct even though they are outliers, as these can be the most interesting and informative examples in a dataset. We term these examples \emph{unique}.
The problems of detecting and leveraging the unique examples in a dataset has received less attention, and the work that does exist focuses on identifying complete documents or segments of documents that are outliers out of a broader set of documents:
\citet{Guthrie:2007} used manually defined feature vectors to identify segments of documents with anomalous style, topic, or tone, and
\citet{Kumaraswamy:2015} and \citet{Zhuang:2017} construct statistical models, identifying complete documents that are outliers within a set based on semantic variation.

Finally, a related but distinct topic is novelty detection \citep{Soboroff:2005,Lee:2015,Ghosal:2018}, in which two sets of documents are provided, one that is assumed to be known, and one that may contain new content.
The task is to identify novel content in the second set.
While outlier detection methods are often applied to this problem, the inclusion of the known document set makes the task fundamentally different from the problem we consider in this work.

\subsection{Data Collection}
We build on prior work employing online crowd workers to create data by paraphrasing.
In particular, we refine the idea of iteratively asking for paraphrases, where each round prompts workers with sentences from the previous round, leading to more diverse data \citep{Negri:2012,Jiang:2017,Kang:2018}.
We also apply the idea of a multi-stage process, in which a second set of workers check paraphrases to ensure they are correct \citep{Buzek:2010,Burrows:2013,snips}.
Most notably, by incorporating our outlier detection method, we are able to automate detecting detrimental data points while also prompting workers in subsequent rounds to paraphrase more unique examples. 

\section{Outlier Detection}

We propose a new outlier detection approach using continuous representations of sentences.
Using that approach, we explored two applications:
(1) identifying errors in crowdsourced data,
and (2) guiding data collection in an iterative pipeline.

\subsection{Method}

We detect outliers in a dataset as follows:

\begin{enumerate}
\setlength{\itemsep}{0pt}
    \item Generate a vector representation of each instance.
    \item Average vectors to get a mean representation.
    \item Calculate the distance of each instance from the mean.
    \item Rank by distance in ascending order.
    \item (Cut off the list, keeping only the top $k$\% as outliers.)
\end{enumerate}

The final step is parenthesized as in practice we use a dynamic threshold approach, allowing the user to go through as much or as little of the list as they like.

The intuition behind this approach is that we expect our representations to capture the semantic structure of the space for each class.
An example that is far away from other examples in the set is therefore less semantically similar in some sense, making it an outlier.
Importantly, it may be an outlier for two distinct reasons: (1) it is not a valid instance of this class (i.e., an \textit{error}), or (2) it is an unusual example of the class (i.e., \textit{unique}).

This approach is applied independently to each class of data.
As example applications we consider two dialog system  tasks: intent classification and slot-filling.
For classification, data for each possible intent label is considered separately, meaning we find outliers in the data by considering one intent class at a time.
For slot-filling, we group the data into classes based on combinations of slots.

This outlier detection method is rather simple as it relies only on a sentence embedding method, a distance metric, and a threshold $k$;
no other hyperparameters are involved.
Moreover, the method requires no training.
We shall see in Section 4 that this method performs well compared to baseline methods, no matter what sentence embedding method is used.
We use Euclidean distance as our distance metric.\footnote{
We found similar results in experiments with a density-based metric, Local Outlier Factor \citep{Breunig:2000}.
}

\subsubsection{Sentence Representations}

Vector representation of sentences is an active area of research and we leverage the following approaches, each of which has been shown to have state of the art results in different use cases:

\paragraph{Universal Sentence Encoder \citep[USE;][]{USE}}
A Deep Averaging Network method, which averages word embeddings and passes the result through a feedforward network.
The USE is trained using a range of supervised and unsupervised tasks.
    
\paragraph{Smooth Inverse Frequency \citep[SIF;][]{SIF}}
A weighted average of word embeddings, with weights determined by word frequency within a corpus.
We consider word embeddings from GloVe \citep{glove} and ELMo \citep{elmo}.
    
\paragraph{Average}
An unweighted average of word embeddings.
While simple, this approach has been shown to be effective for classification \citep{cnnguide} and other downstream tasks \citep{zhu:2018}.
Again, we consider GloVe and ELMo word embeddings as inputs.

\subsubsection{Model Combination}

In addition to ranked lists produced by using these core sentence embeddings, we also investigated aggregating the ranked lists using the Borda count, a rank aggregation technique that has previously been used for combining web search results \citep{Dwork:2001}.

The Borda count aggregates multiple ranked lists of the same set of items into a single ranked list.
First, points are assigned to each item in each list, with an item in position $i$ in a ranked list of length $N$ receiving $N$--~$i$ points.
Next, points for items are summed across all of the lists.
Finally, the items are ranked by their total number of points, producing a final ranking.

\subsection{Application: Error Detection}

Our proposed use of outlier detection to identify errors requires no further processing.
When used in practice, a user looks through the sorted list of examples, either stopping at a given fraction, or when errors become infrequent enough.

\subsection{Application: Uniqueness-driven Data Collection}\label{section:unique}

One core insight in this work is that outlier detection can be used for more than just finding errors.
The outliers that are not errors are likely to be the most interesting and informative examples in our dataset.
We propose to use these examples to guide data collection in an iterative process, with the goal of yielding more diverse data.

To demonstrate this idea, we developed a novel crowdsourcing pipeline for data collection.
Following prior work in crowdsourcing for dialog \citep{Kang:2018,Jiang:2017}, we ask crowd workers to write paraphrases of seed sentences with known intents and slot values.
This provides linguistic diversity in our data in a way that is easily explained to workers.
For instance, given the seed sentence ``\emph{What is my savings account balance?}" a worker might write ``\emph{How much money do I have in my savings account?}".

\begin{figure*}
    \centering
    \begin{subfigure}[b]{0.32\textwidth}
    \centering
        \includegraphics[trim={1.25cm 0 10cm 0},width=\linewidth]{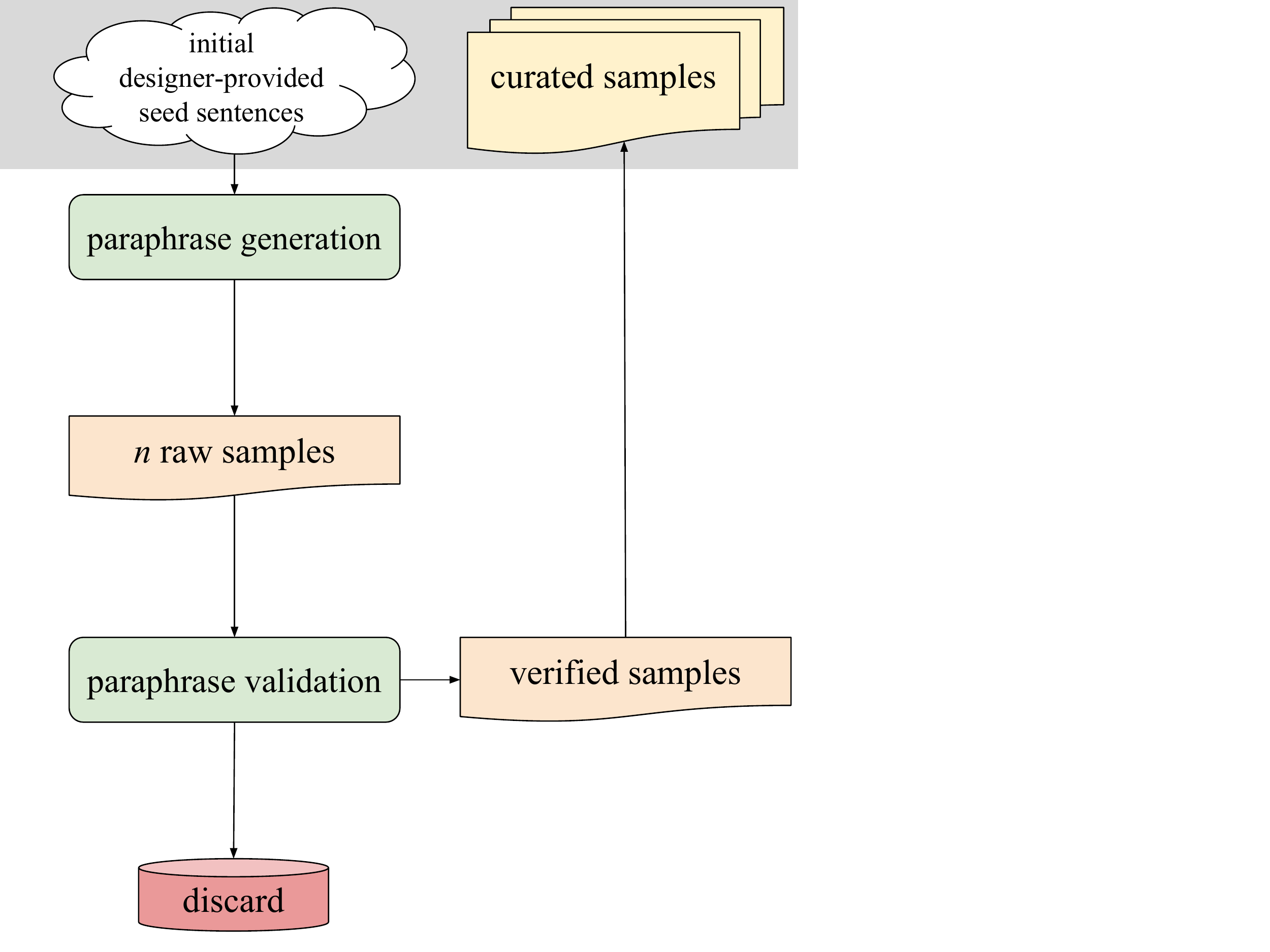}
    \caption{\label{fig:pipeline-old}
    Standard Pipeline}
    \end{subfigure}
    \hspace{2cm}
    \begin{subfigure}[b]{0.32\textwidth}
        \includegraphics[trim={1.25cm 0 10cm 0},width=\linewidth]{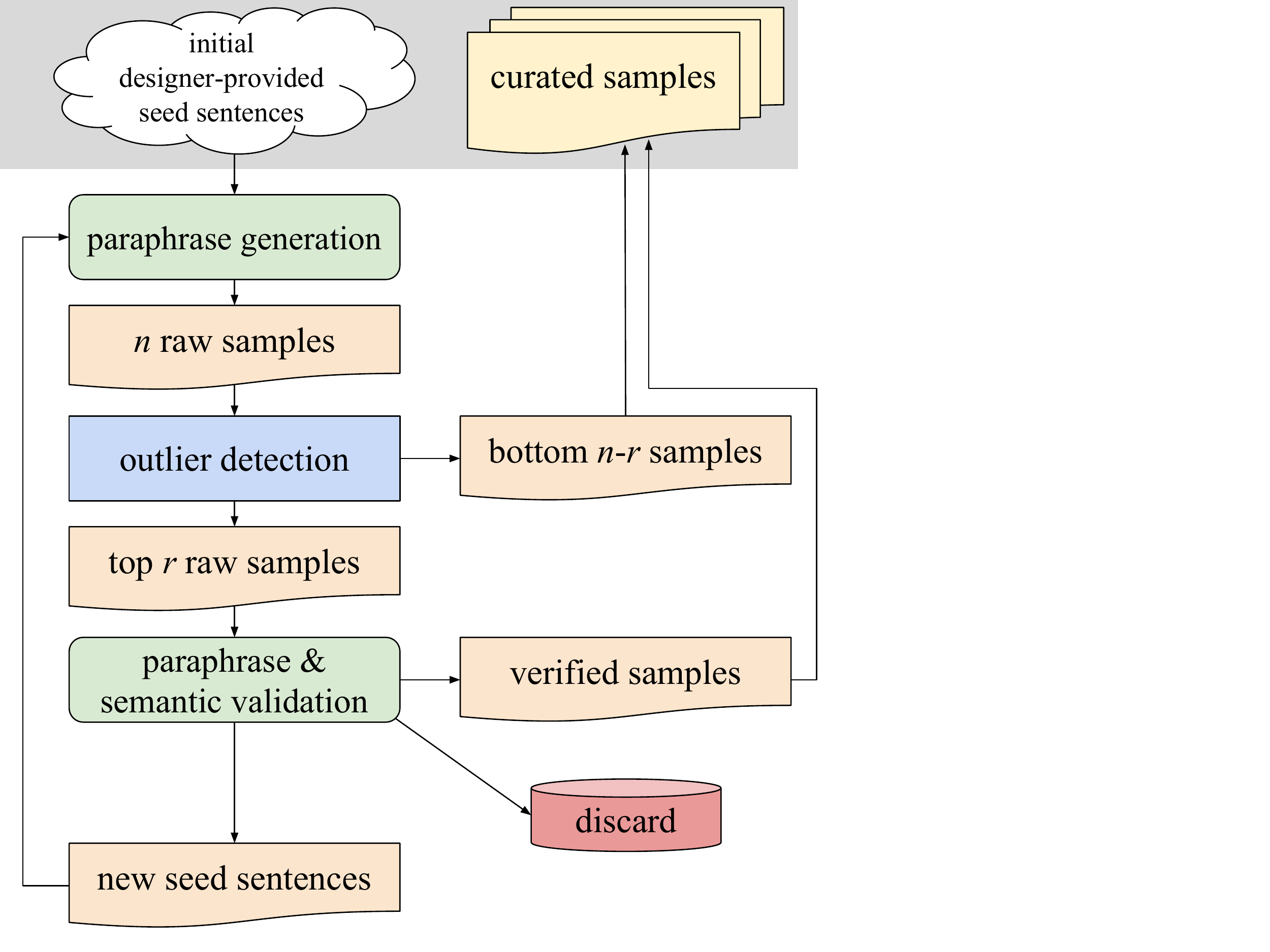}
    \caption{\label{fig:pipeline-new}
    Our Uniqueness-driven Pipeline}
    \end{subfigure}

    \caption{\label{fig:pipeline}
        Data collection pipelines.
        Our outlier detection method (\textcolor{CornflowerBlue}{blue box}) is incorporated into the uniqueness-driven data collection pipeline to guide crowd workers to write more diverse paraphrases.
        \textcolor{YellowGreen}{Green rounded boxes} are manual processes performed by crowd workers, \textcolor{BurntOrange}{orange boxes} with curved bases are data, and the \textcolor{CornflowerBlue}{blue rectangular box} is our outlier detection method.
        In (b), $r$ is the number of outliers detected from $n$ samples.
    }
\end{figure*}

Figure~\ref{fig:pipeline-old} shows a common crowdsourcing pipeline.
The task designer writes seed sentences that target an intent (for classification) or a slot (for slot-filling).
Crowd workers read the seed and write paraphrases.
These paraphrases are then passed to another set of workers who validate if they are in fact accurate paraphrases.

There are two major downsides to this standard pipeline.
First, the validation step increases the cost-per-example.
Second, the diversity of paraphrases depends on the given seed sentences \citep{Jiang:2017}, creating a challenge for the task designer to think creatively.

We introduce a new pipeline, shown in Figure~\ref{fig:pipeline-new} that uses outlier detection to (1) reduce the number of sentences being checked, and (2) collect more diverse examples.
Our new approach uses outlier detection to select only a subset of sentences to be checked: namely, the ones ranked as most likely to be outliers.
This reduces effort by focusing on the sentences most likely to be incorrect.

To try to increase diversity, we also introduce a process with several \emph{rounds} of data collection.
Outlier paraphrases collected in one round are used to seed the next round of data collection.
We could directly use the sentences labeled as correct in the validation step, but while these sentences are correct, they may have diverged from the desired semantics (e.g. diverged from the desired intent class).
To avoid confusion in the next round, we add a step in which workers are shown the most similar sentence from another intent (based on sentence embedding distance) and asked if the new seed is more similar to its intended intent or the alternative example.
Only seeds judged as closer to their intended intent are retained.

This iterative process is intended to collect more diverse data by priming workers to think about ways of phrasing the intent that are not well covered in the current data.
At the same time, we avoid the correctness issues \citet{Jiang:2017} observed by incorporating the validation step.

\section{Experiments}

We perform two sets of experiments to probe the effectiveness of our outlier detection method.
First, we consider error detection, comparing various ranking methods in artificial and real data scenarios.
Second, we use our uniqueness-driven pipeline to collect data, measuring the impact on data diversity and model robustness.
All experiments were conducted on English language data.

\begin{table*}
\centering
\setlength{\tabcolsep}{3.8pt}
    \begin{tabular}{ll|rrrrr|rrrrr}
        \toprule
         & &  & \multicolumn{3}{c}{MAP} & & \multicolumn{5}{c}{Recall@$k$=10\%} \\
        Method & & 1\% & 2\% & 4\% & 8\% & Real & 1\% & 2\% & 4\% & 8\% & Real \\
        \midrule
        \multirow{4}{*}{Baseline} & Random & 0.02 & 0.04 & 0.07 & 0.09 & 0.06 & 0.06 & 0.09 & 0.12 & 0.09 & 0.04 \\
        & Short & 0.02 & 0.05 & 0.07 & 0.11 & 0.27 & 0.12 & 0.13 & 0.12 & 0.15 & 0.27 \\
        & Long & 0.08 & 0.10 & 0.08 & 0.17 & 0.04 & 0.25 & 0.17 & 0.11 & 0.12 & 0.04 \\
        & BoW & 0.14 & 0.14 & 0.14 & 0.24 & 0.10 & 0.41 & 0.24 & 0.24 & 0.22 & 0.10 \\
        \midrule
        \multirow{5}{*}{Neural} & Average GloVe & 0.16 & 0.17 & 0.22 & 0.26 & 0.33 & 0.55 & 0.41 & 0.49 & 0.39 & 0.33 \\
        & Average ELMo & 0.20 & 0.19 & 0.23 & 0.32 & 0.32 & 0.48 & 0.48 & 0.45 & 0.42 & 0.32 \\
        & SIF GloVe (SG) & 0.35 & 0.29 & 0.41 & 0.45 & 0.21 & 0.75 & 0.64 & 0.62 & 0.54 & 0.20 \\
        & SIF ELMo (SE) & 0.30 & 0.37 & 0.48 & 0.52 & 0.20 & 0.87 & 0.74 & 0.70 & 0.65 & 0.19 \\
        & USE (U) & 0.46 & 0.50 & 0.63 & 0.69 & \textbf{0.37} & 0.84 & 0.83 & 0.80 & 0.74 & \textbf{0.38} \\
        \midrule
        \multirow{3}{*}{Combined} & U+SG & 0.51 & 0.54 & 0.66 & 0.68 & 0.36 & 0.84 & 0.81 & 0.83 & 0.76 & 0.33 \\
        & U+SE & \textbf{0.58} & \textbf{0.62} & \textbf{0.68} & \textbf{0.72} & 0.34 & \textbf{0.89} & \textbf{0.87} & \textbf{0.86} & \textbf{0.81} & \textbf{0.38} \\
        & U+SE+SG & 0.51 & 0.57 & 0.67 & 0.69 & 0.36 & 0.87 & 0.84 & 0.84 & 0.78 & 0.34 \\
        \bottomrule
    \end{tabular}
    \caption{\label{tab:map}
    Outlier detection effectiveness for error detection in an artificial setting (constructed by randomly adding content from other intents) and a real setting (manually checked utterances from a crowdsourced set). The artificial results are represented by different values of $p$ (1\%, 2\%, 4\%, and 8\%), where $p$ represents different amounts of errors injected into each intent class.
    Our proposed neural methods are consistently more effective, reducing the manual effort required to identify errors.
    }
\end{table*}

\subsection{Error Detection}

We measure error detection effectiveness in two settings, one artificial and the other more realistic.

\paragraph{Artificial}
First, following prior work, we consider an artificial dataset in which we inject noise by mixing data from different intents \citep{Guthrie:2007,Guthrie:2008,Zhuang:2017,Amiri:2018}.
This provides an easy way to control the amount and type of anomalous data, but does lead to an easier task as the incorrect examples are generally more different than naturally collected errors would be.
The specific data we consider is a set of 20 intents from an in-production dialog system.
To generate outliers for a given intent class $X_i$, we randomly sample $p\cdot |X_i|$ sentences from other intents (e.g. $p=0.04$, or 4\%).

\paragraph{Real}
We collected a new set of sentences for ten intents.
Workers were given three seed sentences per intent and asked to write five paraphrases per seed.\footnote{
Crowd workers were paid 20\textcent~per HIT.
}
Each seed was given to 15 crowd workers, leading to 2250 samples overall, after which duplicates were discarded.
To identify errors, the authors independently checked each sentence and discussed any disagreements to determine a consensus label (either \textit{inlier} or \textit{error}).

\begin{table}
\centering
\scalebox{0.784}{
\setlength{\tabcolsep}{3.8pt}
    \begin{tabular}{ccc}
        \toprule
        Intent & Label & Example \\
        \hline
        & \textit{seed} & what is my withdrawal limit? \\
        withdrawal & \textit{inlier} & how high is my withdrawal ceiling? \\
        & \textit{error} & How much money do I have available \\
        \hline
        & \textit{seed} & what's my balance? \\
        balance & \textit{inlier} & Let me know how much money I have. \\
        & \textit{error} & What can I afford? \\
        \hline
        & \textit{seed} & what's my bank's phone number \\
        phone & \textit{inlier} & I need to call my bank \\
        & \textit{error} & information on my bank \\
        \hline
        & \textit{seed} & i need to order more checks\\
        checks & \textit{inlier} & I need to stock up on more checks \\
        & \textit{error} & No checkbox, more?\\
        \bottomrule
    \end{tabular}
    }
    \caption{\label{tab:examples}
    Examples from the Real dataset. The ``How much money do I have available" example was labeled an error since it is too similar to the balance intent. The ``What can I afford?", ``information on my bank", and ``No checkbox, more?" examples are labeled as errors since they are too vague and ambiguous.
    }
\end{table}

\subsubsection{Evaluation Metrics}

Since our core outlier detection method produces a ranked list, we are interested in evaluating how effective it is at ranking errors near the top.
We use Mean Average Precision (MAP) as an overall measure of list quality.
MAP is the mean over intents of:
\begin{equation*}
    \frac{1}{|\text{errors for intent}|} \sum_{e~\in~\text{errors}} \frac{|\text{errors at or above}~e|}{e}
\end{equation*}
where $e$ is the position of an error in the list.

While this gives an overall qualitative measure for comparison, we are also interested in understanding the precision--recall tradeoff when choosing a threshold $k$ on the ranked lists.
We consider defining the cutoff as a percentage $k$ of the list and measure the percentage of errors that are covered for each possible cutoff.
This measure is equivalent to Recall@$k$, that is,
\begin{equation*}
    \text{Recall@$k$} = \frac{|\text{ errors above $k$}|}{|\text{errors}|}.
\end{equation*}
We average these values across intents to get an overall value for each cutoff percentage $k$.

\subsubsection{Baselines}

For comparison, we consider four simple baselines:
randomly ordering the samples (Random),
sorting from shortest to longest (Short),
sorting from longest to shortest (Long),
and calculating distances in the vector space defined by a bag of words (BoW).

\begin{figure*}
    \centering
    \begin{subfigure}[b]{\textwidth}
    \centering
        \includegraphics[width=\linewidth]{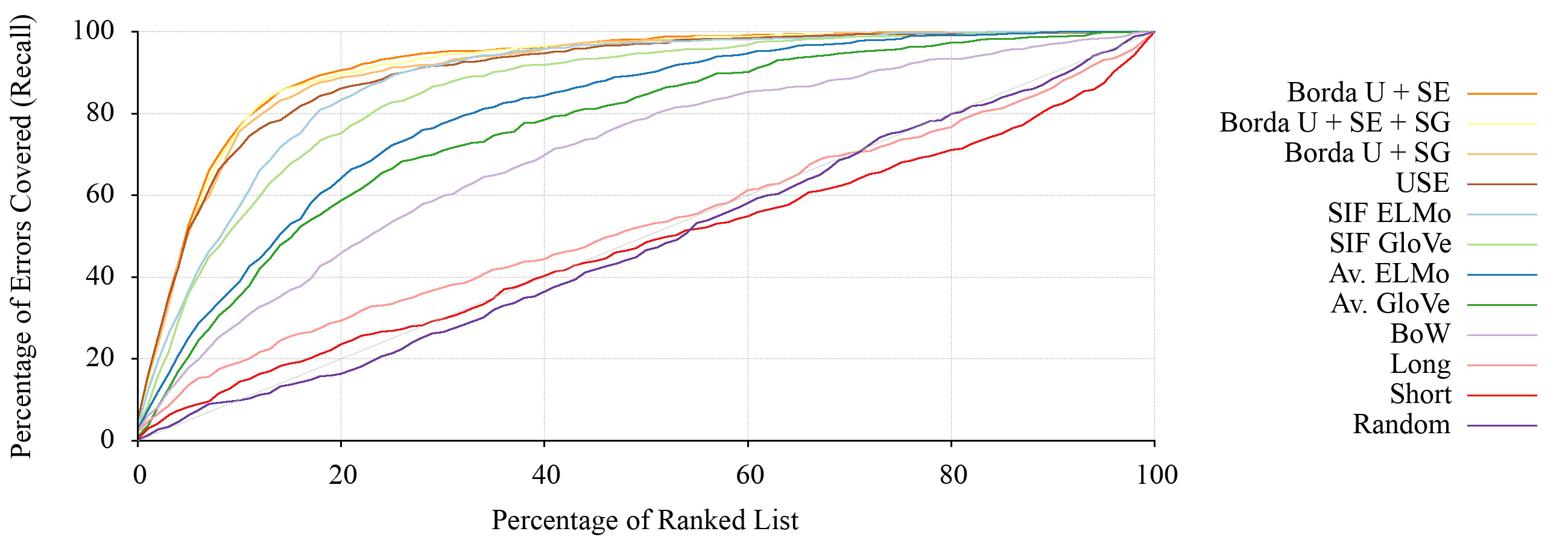}
        \caption{\label{fig:errors-artificial}
        Artificial data.
        }
    \end{subfigure}
    \par\smallskip
    \begin{subfigure}[b]{\textwidth}
    \centering
        \includegraphics[width=\linewidth]{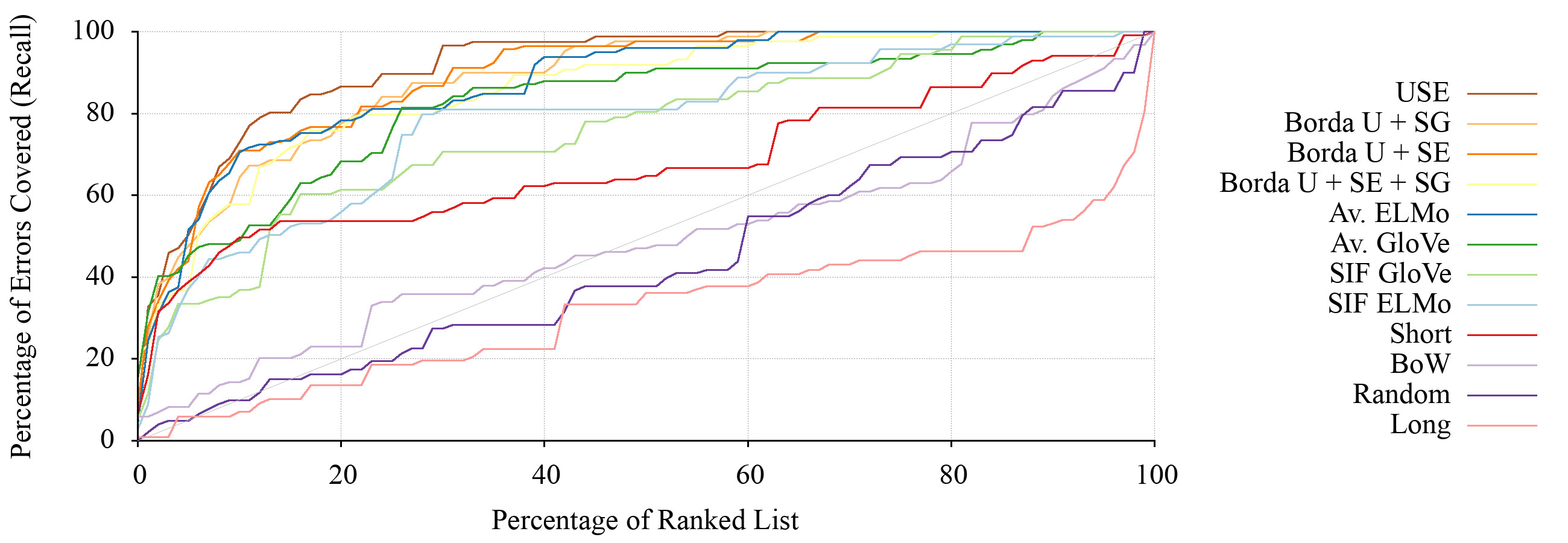}
        \caption{\label{fig:errors-real}
        Real data.
        }
    \end{subfigure}
    \caption{\label{fig:errors}
    Cumulative distribution of errors in ranked lists: a higher line indicates that the method places more errors earlier in the list.
    Results are less smooth for the real data as there are only 51 errors in the set.
    For each plot, the legend is in the same order as the lines at 20\% (i.e., in (a) the top line is Borda U+SE, while in (b) it is USE).
    Neural ranking methods are consistently more effective, with USE covering over 80\% of errors in the first 20\% of the list.
    }
\end{figure*}

\subsubsection{Results}

Table \ref{tab:map} presents MAP and Recall@$k$ for error detection in the two settings (Artificial and Real).
The neural methods outperform the baselines in both settings, demonstrating the effectiveness of our proposed approach.
However, the relative performance of the neural methods differs substantially between the two settings.
Specifically,
(1) SIF performs better than an unweighted average on artificial data, but on real data we see the opposite trend,
(2) combining rankings with Borda appears to help on the artificial data, but not on the real data,
(3) ranking by length is surprisingly effective on the real data,
and (4) results tend to be lower on the real data than the artificial (even at lower values of $p$).
This last point suggests that the commonly used artificial setting does not perfectly capture the types of errors that occur in practice.

For (3), we note that the Short baseline method performs particularly well vis-\`a-vis other baselines on the real data, but not comparatively well on the artificial data.
This can be explained by observing that the length of the average error in the real data is roughly 6 tokens, while the average inlier length is 8 tokens.
Lengths of errors and inliers are roughly the same (roughly 8 tokens) in the artificial dataset, due to the outlier selection scheme.

\begin{figure}
    \fbox{\begin{minipage}{0.96\columnwidth}
    \begin{center}
    \small
    ~\\
    show \hl{average} exchange rate from \sethlcolor{orange}\hl{ten} \sethlcolor{green}\hl{usd} to \sethlcolor{green}\hl{cad} \sethlcolor{pink}\hl{last year}
    ~\\
    \end{center}
    \end{minipage}}
    \caption{\label{fig:svp_slots}
    Example annotated sentence for the slot-filling task. The slot names are (in order of appearance) \sethlcolor{yellow}\hl{\textit{metric}}, \sethlcolor{orange}\hl{\textit{amount}}, \sethlcolor{green}\hl{\textit{currency}}, and \sethlcolor{pink}\hl{\textit{date}}.}
\end{figure}

\begin{figure}
    \centering
    \scalebox{0.75}{\fbox{\begin{minipage}{\columnwidth}
        \includegraphics[width=\linewidth]{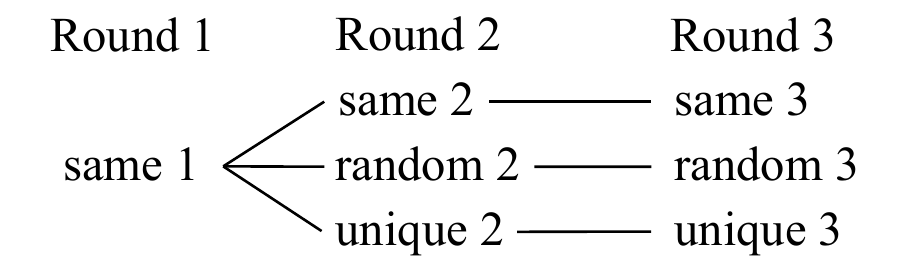}
    \end{minipage}}}
    \caption{\label{fig:rounds}
    Data collection rounds.
    The final datasets combine data from all three rounds along each path.
    }
\end{figure}

While the values in Table \ref{tab:map} allow an overall comparison of the methods, they do not provide a clear qualitative sense of the distribution of errors in the lists.
Figure~\ref{fig:errors} shows the distribution for each method in the two settings.
The effectiveness of the neural methods, and USE in particular, is again clear.
In the real data, when considering just the first 20\% of the list, USE covers over 85\% of the errors on average.
One easy example was ``\emph{No checkbox, more?}" when the intent was to order more checks.
This is clearly an error, which would at the very least need to have \emph{checkbox} replaced by \emph{checkbook}.
In contrast, one hard example for USE was ``\emph{How much money do my banks}" when the intent was to request the user's balance.
Until the last word, this example looks like it will be a valid balance request.
These examples show that the system is qualitatively fitting our expectations for error detection.

\subsection{Uniqueness-driven Data Collection}

The second set of experiments evaluates our proposed uniqueness-driven data collection pipeline.
We consider collecting data for two tasks used by dialog systems: intent classification and slot-filling.
In each case, we calculate intrinsic measures of data diversity and the robustness of models trained on the data.

\paragraph{Tasks}
We consider intent classification with 10 intents related to banking, and slot-filling for foreign exchange rate requests with four slots: \textit{amount}, \textit{currency}, \textit{date}, and \textit{metric}.
Figure~\ref{fig:svp_slots} shows an example query with annotated slots.

\paragraph{Approaches}
As well as our proposed data collection pipeline (\texttt{unique}), we consider a variant where the next seed is chosen randomly (\texttt{random}), and one where the seeds are the same in every round (\texttt{same}).
The third case is equivalent to the standard pipeline from Figure~\ref{fig:pipeline-old}.
All three pipelines start from the same first round and then vary in the subsequent rounds, as depicted in Figure~\ref{fig:rounds}. Each pipeline collected data for three rounds.
The final dataset for each approach combines data collected from all three rounds.

In both tasks, we asked workers to rephrase each seed sentence 5 times and showed each seed sentence to 15 workers.
For classification there were 3 seed sentences per intent.
For slot-filling we defined 4 example scenarios, each corresponding to a specific combination of slots. We used Borda USE+SG with $k$ set to 10\% for the outlier detection model.

\subsubsection{Evaluation Metrics}

We consider several different metrics to probe how effectively our proposed pipeline improves data quality.
In all cases, higher values are better.

\paragraph{Intrinsic}
We measure the diversity and coverage of each dataset using the metrics introduced in \citep{Kang:2018} and shown in Figure~\ref{fig:metrics}.
%\footnote{
%Note, the scale factor of $\frac{1}{N}$ in $D(a, b)$ is missing from the equation in their paper (verified via author correspondence), but is needed to calculate the mean over n-gram lengths as described in the text.
%}

\paragraph{Extrinsic}
The main reason to increase dataset diversity is to construct more robust models.
To directly evaluate that objective, we randomly divided the datasets collected by each pipeline into training and test sets (85-15 split).
Our intuition is that a robust model should perform fairly well across all test sets.
Training on a dataset that is not diverse will lead to a brittle model that only does well on data collected with the same seed sentences.
For intent classification, we measure accuracy of two models:
an SVM \citep{SVM} using bag of words feature representation, and FastText \citep{FastText}, a neural network that averages across sentence embeddings and passes the result through feedforward layers.
For slot-filling, we measure the F$_1$-score of a bi-directional LSTM with word vectors that are trained, but initialized with GloVe 300-dimensional embeddings.
For all models, we average results across 10 runs.

\begin{figure}
    \small
    \begin{equation*}
        D(a,b) = 1 - \frac{1}{N} \sum_{n=1}^{N} \frac{| n\mbox{-}\text{grams}_a \cap n\mbox{-}\text{grams}_b |}{| n\mbox{-}\text{grams}_a \cup n\mbox{-}\text{grams}_b |}
    \end{equation*}
    \begin{equation*}
        \text{Diversity}(X) = \frac{1}{|I|} \sum_{i=1}^{|I|} \frac{1}{|X_i|^2} \Bigg[ \mathop{\sum_{a \in X_i} \sum_{b \in X_i}} D(a,b) \Bigg]
    \end{equation*}
    \begin{equation*}
        \text{Coverage}(X,Y) = \frac{1}{|I|} \sum_{i=1}^{|I|} \frac{1}{|Y_i|} \sum_{b\in Y_i} \max_{a\in X_i} (1 - D(a,b))
    \end{equation*}
    \caption{\label{fig:metrics}
    Metrics for diversity and coverage from \citet{Kang:2018}.
    $X$ and $Y$ are sets of utterances labeled with intents from $I$, and $X_i$ is the data in $X$ for intent $i$.
    The distance metric for comparing a pair of utterances is based on the Jaccard Index over \textit{n}-grams.
    We follow the work from \citet{Kang:2018} and set $N$, the max \textit{n}-gram length, to $3$.
    }
\end{figure}

\begin{table}
\centering
    \begin{tabular}{lrrrr}
        \toprule
        & \multicolumn{3}{r}{Data Collection Round} & \\
                 & 1        & 2 & 3 & All \\
        \midrule
        \multicolumn{5}{l}{Diversity} \\
        \texttt{same}      & 0.907    & 0.913 & 0.902 & 0.911    \\
        \texttt{random}       & --     & 0.908 & 0.912 & 0.916     \\
        \texttt{unique}       & --     & \textbf{0.932} & \textbf{0.941} & \textbf{0.944}    \\
        \midrule
        \multicolumn{5}{l}{Samples} \\
        \texttt{same} & 2040 & 2025 & 2024 & 5648 \\
        \texttt{random} & -- & 2010 & 2007 & 5747 \\
        \texttt{unique} & -- & 2083 & 1954 & 5922 \\
        \bottomrule
    \end{tabular}
    \caption{\label{tab:clf_diversity}
    Classification:
    Diversity scores for data collected in each round (top), and the number of samples collected (bottom).
    The data for the All column combines the previous two sets in the row and the data from (\texttt{same} round 1).
    The \texttt{unique} approach produces data that is considerably higher diversity.
    }
\end{table}

\begin{table}
\centering
\setlength{\tabcolsep}{3pt}
    \begin{tabular}{llcc>{\columncolor[gray]{0.85}}c}
        \toprule
         & & \multicolumn{3}{c}{Test Set} \\
        Metric & Training & \texttt{same} & \texttt{random} & \texttt{unique} \\
        \hline
        \multirowcell{3}{SVM \\ Accuracy} & \texttt{same} & \textbf{0.99} & 0.97 & 0.81 \\
        & \texttt{random} & 0.98 & \textbf{0.98} & 0.81 \\
        & \texttt{unique} & 0.99 & 0.97 & \textbf{0.98} \\
        \hline
        \multirowcell{3}{FastText \\ Accuracy} & \texttt{same} & \textbf{0.98} & 0.97 & 0.80 \\
        & \texttt{random} & 0.98 & \textbf{0.99} & 0.83 \\
         & \texttt{unique} & 0.98 & 0.98 & \textbf{0.98} \\
        \hline
         & \texttt{same}   & \textbf{0.68} & 0.64 & 0.45 \\
        Coverage & \texttt{random} & 0.67 & \textbf{0.66} & 0.44 \\
        & \texttt{unique} & 0.64 & 0.58 & \textbf{0.56} \\
        \bottomrule
    \end{tabular}
    \caption{\label{tab:clf_accuracy}
    Classifier accuracy when training on one dataset and testing on another (top and middle), and coverage of the test set for each training set (bottom).
    As expected, the highest scores are when we train and test on the same data, but off the diagonal the \texttt{unique} test set (gray column) is considerably harder for models trained on other data while a model trained on \texttt{unique} performs consistently well.
    This accuracy trend is matched in coverage.
    }
\end{table}

\begin{table}
\centering
    \begin{tabular}{lrrrr}
        \toprule
        & \multicolumn{3}{r}{Data Collection Round} & \\
                 & 1 & 2 & 3 & All \\
        \midrule
        \multicolumn{5}{l}{Diversity} \\
        \texttt{same}      & 0.916      & 0.911 & 0.893 & 0.909     \\
        \texttt{random}    & --     & 0.913 & 0.910 & 0.915     \\
        \texttt{unique}    & --     & \textbf{0.926} & \textbf{0.935} & \textbf{0.930}    \\
        \midrule
        \multicolumn{5}{l}{Samples} \\
        \texttt{same} & 994 & 911 & 952 & 2717 \\
        \texttt{random} & -- & 941 & 923 & 2808 \\
        \texttt{unique} & -- & 977 & 988 & 2914 \\
        \bottomrule
    \end{tabular}
    \caption{\label{tab:svp_diversity}
    Slot-filling:
    Diversity scores for data collected in each round (top), and the number of samples collected (bottom).
    The data for the All column combines the previous two sets in the row and the data from (\texttt{same} Round 1).
    As seen for intent classification, the \texttt{unique} approach produces data that is of considerably higher diversity.
    }
\end{table}

\begin{table}
\centering
\setlength{\tabcolsep}{3pt}
    \begin{tabular}{llcc>{\columncolor[gray]{0.85}}c}
        \toprule
        & & \multicolumn{3}{c}{Test Set} \\
        Metric & Training & \texttt{same} & \texttt{random} & \texttt{unique} \\
        \hline
        \multirowcell{3}{Slot \\ F$_1$} & \texttt{same} & 96.4 & 96.0 & 93.1 \\
        & \texttt{random} & 96.4 & \textbf{96.8} & 93.6 \\
        & \texttt{unique} & \textbf{96.7} & 96.5 & \textbf{94.9} \\
        \hline
        & \texttt{same}   & \textbf{0.812} & 0.788 & 0.726 \\
        Coverage & \texttt{random} & 0.736 & \textbf{0.764} & 0.660 \\
        & \texttt{unique} & 0.761 & 0.752 & \textbf{0.774} \\
        \bottomrule
    \end{tabular}
    \caption{\label{tab:svp_f1_scores}
    F$_1$-scores and coverage scores for each train-test pair for the slot-filling experiment.
    Training on the \texttt{unique} data produces a more robust model, with consistently high performance across test sets.
    }
\end{table}

\begin{table*}
\centering
\scalebox{0.7}{
    \begin{tabular}{ ccccc }
    \toprule
    Intent & Pipeline & Round 2 & Round 3 \\
    \midrule
    \multirow{6}{*}{routing} & \multirow{3}{*}{\texttt{random}} & Where can I find the routing number for my bank? & what is the best routing number for my bank \\
    & & Do you have my bank's routing number? & can you help find my bank routing number? \\
    & & show me my banks routing number & I need to see the routing number for my bank \\
    \cmidrule(r){2-4}
    & \multirow{3}{4em}{\texttt{unique}} & acquire my banks routing number & what is the correct ABA? \\
    & & how does a person find their correct routing number? & How do I find the ABA? \\
    & & I'm looking for the router number for my bank. & Whats the router number? \\
    \midrule
    \multirow{6}{*}{hours} & \multirow{3}{*}{\texttt{random}} & Can you tell me when my bank is open? & How long is my bank open? \\
    & & When does my bank open? & What are the hours for my bank?  \\
    & & What time is the bank open & What time will the bank be open \\
    \cmidrule(r){2-4}
    & \multirow{3}{*}{\texttt{unique}} & display when the bank closes & What hours do you carry \\
    & & What is the earliest you are open? & what are your operating hours? \\
    & & look up the hours of operation for my bank & What is the latest I can come in to a physical branch? \\
    \midrule
    \multirow{6}{*}{checks} & \multirow{3}{*}{\texttt{random}} & i require ordering more checks. & I'd like additional checks \\
    & & Get me more checks. & Can you get me more checks?  \\
    & & can you explain to me how to order additional checks & Please order more checks for me. \\
    \cmidrule(r){2-4}
    & \multirow{3}{*}{\texttt{unique}} & I need to stock up on more checks & what is check ordering procedure? \\
    & & in what manner would i get more checks & what is the fastest method to order checks? \\
    & & Could you rush me some more checks? I'm nearly out. & Teach me how to get more checks. \\
    \bottomrule
    \end{tabular}
    }
    \caption{\label{tab:seeds}
    Seed sentences for selected intents for the classification task.
    The \texttt{unique} approach leads to changes like the use of \emph{ABA} for \emph{routing} (top), and grammatical variations in the sentences for requesting checks (bottom). Examples of seeds for Round 1 include ``what is my bank's routing number?", ``when does the bank close?", and ``how do i order more checks?".
    }
\end{table*}

\subsubsection{Results}

\paragraph{Classification}
Table~\ref{tab:clf_diversity} presents the number of examples and diversity of data collected in each round with each approach.
Diversity is consistently higher with seeds chosen using our proposed \texttt{unique} approach.
Dataset sizes vary because of the removal of duplicates. The \texttt{unique} approach produces a larger final set as there is less duplication across rounds.

Table~\ref{tab:clf_accuracy} displays accuracy scores and coverage for each combination of train and test sets.
As expected, the highest scores are on the diagonal---training and testing on the same source data.
More importantly however, training on the \texttt{unique} data produces a model that is robust, performing well across all three test sets.
In contrast, training on the \texttt{same} or \texttt{random} data produces models that perform substantially worse on the \texttt{unique} test set.
This trend is also present in the coverage scores in the bottom section of the table.

Table~\ref{tab:seeds} shows some of the seed sentences produced by the \texttt{unique} and \texttt{random} approaches.
These examples illustrate the trends in our metrics, with the seeds for the \texttt{random} approach often being very similar.
Meanwhile, the \texttt{unique} approach produces seeds with grammatical variation and the introduction of quite different expressions, such as ``ABA" instead of ``routing number".

\paragraph{Slot-filling}
Table~\ref{tab:svp_diversity} shows the number of samples collected per round for each of the data collection pipelines and the diversity of the sets.
As in the classifier experiment, we observe that data produced by the \texttt{unique} pipeline is of higher diversity than the other two pipelines.

Table~\ref{tab:svp_f1_scores} displays F$_1$-scores and coverage for each train--test combination.
Again, we see the same trends, with training on \texttt{same} or \texttt{random} leading to low results on the \texttt{unique} dataset, but not the reverse, and similarly for coverage, though the gaps are smaller than for classification.

\section{Conclusion}

Outliers are often the most interesting parts of our data, but outlier detection has received relatively little attention in NLP beyond its application to finding annotation errors.
This paper introduces the first neural outlier detection method for short text and demonstrates its effectiveness across multiple metrics in multiple experiments.

We also propose a way to integrate outlier detection into data collection, developing and evaluating a novel crowdsourcing pipeline.
This pipeline supports the creation of higher quality datasets to yield higher quality models by both reducing the number of errors and increasing the diversity of collected data. 
While the experiments discussed herein are concerned with components of dialog systems, we believe that similar data collection strategies could yield benefits to other areas of NLP as well.

\section*{Acknowledgments}
The authors thank Yiping Kang, Yunqi Zhang, Joseph Peper, and the anonymous reviewers for their helpful comments and feedback.

\balance{

}
\end{document}